\documentclass[conference]{IEEEtran}
\IEEEoverridecommandlockouts
\usepackage{cite}
\usepackage{amsmath,amssymb,amsfonts}
\usepackage{algorithmic}
\usepackage{graphicx}
\usepackage{textcomp}
\usepackage{xcolor}
\def\BibTeX{{\rm B\kern-.05em{\sc i\kern-.025em b}\kern-.08em
    T\kern-.1667em\lower.7ex\hbox{E}\kern-.125emX}}

\begin{document}

\title{Long-Term Planning Around Humans \\ in Domestic Environments with 3D Scene Graphs
}

\author{
    \IEEEauthorblockN{1\textsuperscript{st} Ermanno Bartoli}
    \IEEEauthorblockA{\textit{KTH Royal Institute of Technology} \\
    bartoli@kth.se}
    \and
    \IEEEauthorblockN{2\textsuperscript{nd} Dennis Rotondi}
    \IEEEauthorblockA{\textit{University of Stuttgart} \\
    dennis.rotondi@ki.uni-stuttgart.de}
    \and[\hfill\mbox{}\par\mbox{}\hfill]
    \IEEEauthorblockN{3\textsuperscript{rd} Kai O. Arras}
    \IEEEauthorblockA{\textit{University of Stuttgart} \\
    kai.arras@ki.uni-stuttgart.de}
    \and
    \IEEEauthorblockN{4\textsuperscript{th} Iolanda Leite}
    \IEEEauthorblockA{\textit{KTH Royal Institute of Technology} \\
    iolanda@kth.se}
}
\maketitle

\begin{abstract}
Long-term planning for robots operating in domestic environments poses unique challenges due to the interactions between humans, objects, and spaces. Recent advancements in trajectory planning have leveraged vision-language models (VLMs) to extract contextual information for robots operating in real-world environments. While these methods achieve satisfying performance, they do not explicitly model human activities. Such activities influence surrounding objects and reshape spatial constraints. This paper presents a novel approach to trajectory planning that integrates human preferences, activities, and spatial context through an enriched 3D scene graph (3DSG) representation. By incorporating activity-based relationships, our method captures the spatial impact of human actions, leading to more context-sensitive trajectory adaptation. Preliminary results demonstrate that our approach effectively assigns costs to spaces influenced by human activities, ensuring that the robot’s trajectory remains contextually appropriate and sensitive to the ongoing environment. This balance between task efficiency and social appropriateness enhances context-aware human-robot interactions in domestic settings. Future work includes implementing a full planning pipeline and conducting user studies to evaluate trajectory acceptability.
\end{abstract}


\section{Introduction}
Long-term Human-Robot Interaction (HRI) aims to create robots that continuously adapt their behavior by learning from ongoing interactions with humans \cite{shaheen2022continual}. This capability is essential for assistive robots operating in domestic environments, where they must not only execute tasks but do so in a way that is sensitive to human preferences. Effective robot behavior in these settings requires understanding not just spatial constraints but also the activities humans engage in and how these activities influence the environment.

A fundamental challenge in such environments is motion planning, which extends beyond simple obstacle avoidance to ensuring socially appropriate navigation. Robots must account for human activities and preferences, making decisions that respect explicit instructions, such as ``watch out for the glass table; it could break," and implicit contextual cues, such as ``don't spill the glass of wine," suggesting caution in the navigation. For example, a robot performing a cleaning task should avoid obstructing the human's line of sight while watching television or interfering with their ongoing activities. Achieving this requires a fine-grained understanding of human behavior and its spatial implications.

This complexity is further amplified by static objects in the environment, whose significance changes dynamically based on human engagement. 
For example, if a person watches TV, the space between them and the screen becomes socially significant, requiring the robot to avoid passing through it unless necessary. Similarly, an armchair that is currently unoccupied differs in relevance from one actively used by a human. 
These cases suggest the need for an approach that evaluates an object's relevance based on the human's current activities and involvement with it.

\begin{figure}[t!]
\centering
\begin{minipage}{0.49\textwidth}
  \centering
  \includegraphics[width=0.80\linewidth]{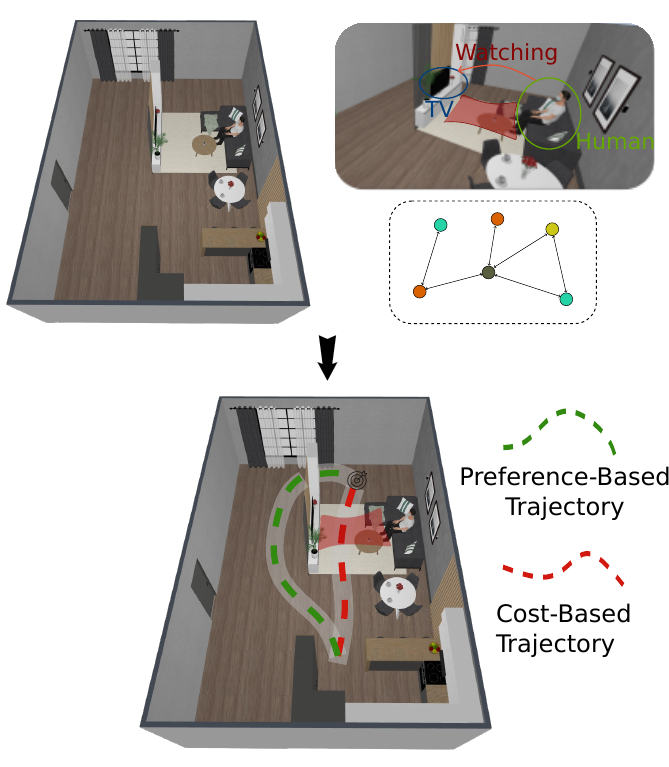}
    \caption{Overview of the method. Starting from a 3D map and its 3D scene graph representation, our approach computes a preferred based trajectory which is socially aware of the human presence in the scene.}
    \label{fig:trajs}
\end{minipage}
\end{figure}

In the context of social navigation, this presents several challenges: balancing task efficiency with social appropriateness, adapting to dynamic human behaviors, and ensuring long-term generalization. Traditional motion planning in human-shared environments primarily focused on respecting personal space, but respecting social norms is equally important \cite{kruse2013human, rudenko2020human}.
Moreover, as human activities continuously reshape the environment, static objects take on varying importance \cite{luber2012socially, sathyamoorthy2021comet}, requiring robots to interpret these dynamics accurately. Learning implicit human preferences over time adds another layer of complexity, necessitating structured representations that can store and infer relevant information \cite{jeon2020rewardrationalimplicitchoiceunifying, Holk_2024}. A key challenge remains generalization, as models trained in controlled settings often fail to transfer effectively to real-world scenarios \cite{pfeiffer2016predicting, schilling2017geometric}.

To address these issues, we propose an approach that integrates human preferences, activities, and spatial configurations into a 3D scene graph representation. Unlike purely vision-language-based models, which provide rich semantic understanding but lack structured spatial reasoning, 3D scene graphs enable both real-time interpretation and long-term planning. By capturing human activities as graph relationships, robots can dynamically adjust their motion plans based on human presence, ensuring socially aware navigation that respects both the immediate and evolving context. This structured yet adaptable representation allows for more intuitive and personalized human-robot interactions in shared environments, bridging the gap between spatial reasoning and socially intelligent behavior.

\begin{figure*}[t!]
\centering
\begin{minipage}{0.99\textwidth}
  \centering
  \includegraphics[width=\linewidth]{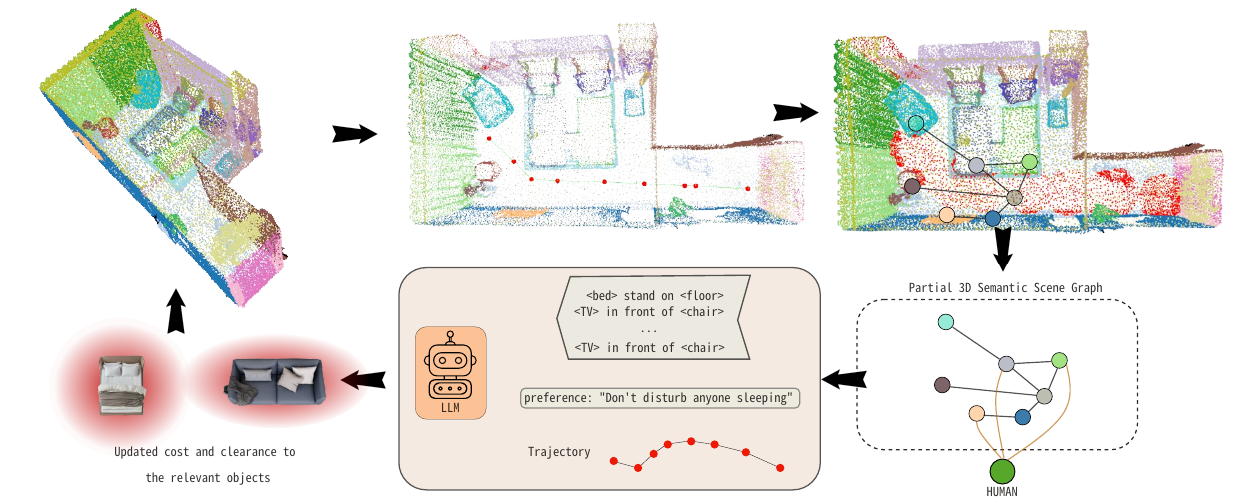}
  \caption{Our proposed approach constructs an object-centric description of impact factors, considering both objects and spaces, through a human-centered investigation of the scene. Starting from a 3D map, a trajectory, and a set of preferences, we: (a) extract the Partial 3D Semantic scene graph with objects that could potentially impact the trajectory; (b) enrich the graph by incorporating the human and their activities, integrating with existing nodes into the 3D scene graph; and (c) feed the enriched graph representation, along with the trajectory and preferences, into a Large Language Model (LLM). The LLM then calculates for each object of interest a cost, combined with a clearance value, that describes how the cost decreases with distance. }
  \label{fig:images/pipeline1.pdf}
\end{minipage}%
\end{figure*}

\section{Related Work}
\subsection{3D Scene Graph Resources}
3D scene graphs (3DSGs) \cite{Kim_2020, armeni20193dscenegraphstructure} are designed for robotics, featuring a hierarchical structure where nodes represent scene parts like rooms and floors. At the lowest level, objects are connected by spatial (e.g., object1 \textit{is next to} object2) and comparative relationships (e.g., object1 \textit{is larger than} object2).
Currently, only two datasets include 3D scene graphs \cite{armeni20193dscenegraphstructure, wald2020learning3dsemanticscene}, focusing on standardizing object classes via WordNet \cite{wordnet}. However, they don't incorporate active base relationships, which are present in semantic inventories \cite{propbank, verbatlas, nounatlas}. These datasets are static, lacking human presence, which limits robotic applications in dynamic, human-populated environments.
To address this, 3D Dynamic Scene Graphs \cite{rosinol2021kimeraslamspatialperception, rosinol20203ddynamicscenegraphs, gorlo2024longtermhumantrajectoryprediction, Honerkamp_2024, Greve_2024} have been introduced to account for dynamic scenes and agents like humans. However, the focus has mostly been on tasks like trajectory prediction and autonomous driving, with limited exploration of reasoning over dynamic 3D scene graphs.



\subsection{Planning with 3D Scene Graphs}
The 3D scene graph structure, often built on SLAM or processed images and point clouds, is as powerful as any pre-learned representation, making it ideal for robotics applications like localization \cite{conceptgraph, miao2024scenegraphloccrossmodalcoarsevisual}, navigation \cite{ravichandran2022hierarchicalrepresentationsexplicitmemory, yin2024sgnavonline3dscene, hov-sg}, and planning \cite{liu2024deltadecomposedefficientlongterm, agia2022taskographyevaluatingrobottask, rana2023sayplangroundinglargelanguage, dai2024optimalscenegraphplanning, ray2024taskmotionplanninghierarchical}.
Planning typically involves using a large language model (LLM) to interpret a text-based 3D scene graph with object positions, bounding boxes, labels, and relationships. This helps robots understand goals and plan tasks, such as moving objects to reach their goal.
A less explored but important area is incorporating object affordances and attributes into decision-making. For example, this could guide navigation to avoid fragile objects or stepping on valuable surfaces like carpets or clothes.
Current 3DSG planners do not account for dynamic agents and their interactions with the environment, limiting their use to static scenes. This is suitable for environments with minimal human involvement but inadequate for dynamic settings like homes or offices.
\subsection{Social Navigation}

Social navigation is a broad research area that has evolved significantly over the years. Early works primarily focused on learning the social use of space through the lens of proxemics, studying how humans naturally maintain spatial boundaries and personal space \cite{mead2017autonomous, mumm2011human, takayama2009influences}. A key advancement in the field has been the incorporation of richer environmental representations, which allow robots to better understand and navigate spaces in a socially acceptable manner \cite{kruse2013human, rudenko2020human}. Approaches to social navigation have been developed using both supervised and unsupervised learning techniques \cite{luber2012socially}. The problem has been explored in both outdoor \cite{schilling2017geometric} and indoor \cite{pfeiffer2016predicting} settings, with a particular focus on crowded spaces \cite{sathyamoorthy2021comet} and human-robot encounters \cite{10341993}. However, a persistent challenge across these approaches has been generalization, ensuring that learned behaviors remain effective across diverse environments and interactions.

With the advent of LLMs and vision-language models (VLMs), significant progress has been made toward richer semantic understanding, enabling more complex robot behaviors informed by contextual cues \cite{firoozi2023foundation}. Early works identify navigation targets \cite{shah2023lm}, and recent advancements have focused on using them to guide low-level navigation behaviors, ensuring that robots adapt their movements in a socially appropriate manner based on the specific scenario \cite{sathyamoorthy2024convoi}. 


While LLMs and VLMs enhance contextual understanding by capturing human presence and activities, they lack an explicit, structured representation that supports long-term planning. We foresee that this structure can be provided by extending 3DSG representations to include humans and relate them to their surroundings.



\section{Proposed Methodology}

\subsection{Preliminary Considerations}
We consider a scene represented by a 3DSSG as provided in \cite{wald2020learning3dsemanticscene}, where nodes correspond to objects in the environment and edges define spatial relationships between them ( e.g., "on top of", "next to", "hanging on"). However, existing 3DSSG datasets lack human presence. As a preliminary step, we manually introduce one or more humans into the scene and integrate them into the scene graph based on their chosen activities. This is done by creating a new node labeled "human" and associating it with relevant objects through both spatial and activity-based relationships. Spatial relationships (e.g., "sitting on", "standing next to") define the human's physical interaction with the environment, while activity-based relationships (e.g., "reading", "watching", "speaking") capture contextual interactions with objects. The result is a 3DSSG that integrates human presence.

\subsection{Planning Appropriate Trajectories}
We define the problem of trajectory planning in a shared human-robot environment as a scenario where the robot must navigate from a starting point (A) to a goal point (B) while accounting for human activities that may impact the scene. Additionally, the robot may be provided with explicit or implicit preferences that it must incorporate during the planning process.

Given the 3DSSG, along with the robot's task, starting and goal points, and relevant preferences, the robot's objective is to plan a trajectory that aligns with these preferences and is appropriate for the human presence and activities within the environment.

Given a trajectory, which consists of a set of waypoints described by $T = \{ p_1, p_2, p_3, \dots, p_n \} \quad \text{for} \quad i = 1, 2, 3, \dots, n$ where \( n \) is the total number of waypoints along the trajectory, we investigate the potential objects that could impact the trajectory. This is done by searching within a defined radius around each waypoint along the trajectory, which can be adjusted as needed, as shown in Fig. \ref{fig:images/pipeline1.pdf}. Once the relevant objects are identified, they are processed, and the following description is generated for each object. Each object is represented by:

\begin{itemize}
  \item \texttt{``object\_id``}: the id of the object
  \item \texttt{``object\_tag``}: the label of the object
  \item \texttt{``bbox\_center``}: centroid of the 3D bounding box for the object
  \item \texttt{``bbox\_extent``}:  extents of the 3D bounding box for the object
  \item \texttt{``affordances``}: The set of the affordances of the object
  \item \texttt{``attibutes``}: The set of the attributes of the object
  \item \texttt{``relations``}: a set of tuples (\texttt{name of relation}, \texttt{tail entity})
\end{itemize}

Consequently, the Partial 3DSSG, enriched with the human's information, the trajectory and the preferences are inputted to an LLM which is responsible to return for each relevant object a cost and a clearance.
The cost is a value equal or greater than 1 (1 if no impact at all), and reflects the impact factor of the object in the trajectory, while the clearance is a value equal or greater than 0 ( 0 if no impact at all) and acts as a diminishing factor for the cost, reducing the impact as the robot moves farther from the object. This ensures that the robot adjusts its trajectory based on both the proximity to objects and human preferences, maintaining safety and efficiency.

The computed costs and clearances can be integrated into a cost-based planner to generate an optimal, yet human-aware trajectory. Since these values reflect activity-based influences and spatial relationships, the resulting trajectory would inherently respect human presence and preferences. Our approach focuses on cost assignment via an LLM, while planning remains an extension.

\section{Preliminary Findings}

We evaluated our approach using the scene shown in Fig. \ref{fig:images/pipeline1.pdf}, where a human was manually placed in the scene, sitting on the bed and watching TV. The costs of the relevant nodes along the trajectory were extracted, focusing on the following objects: {bed, human, armchair}. We compared our approach—incorporating human information and both spatial and activity-based relationships—against two baseline planners: one using the 3DSSG without human information, and another using the 3DSSG with human information but excluding activity-based relationships. The preliminary results are presented in Table \ref{tab:tab1}.

\begin{table}[htbp]
\caption{Table Type Styles}
\begin{center}
\begin{tabular}{|c|c|c|c|}
\hline
\textbf{}&\multicolumn{3}{|c|}{\textbf{Cost (Clearance)$^{\mathrm{*}}$}} \\
\cline{2-4} 
\textbf{} & \textbf{\textit{Bed}}& \textbf{\textit{Human}}& \textbf{\textit{armchair}} \\
\hline
No Human& 1 (0.5) & - & 2 (1.5) \\
\hline
Human w/out relations& 2 (0.5) & 10 (2) & 3 (1) \\
\hline
\textbf{Human$^{\mathrm{**}}$ w/ relations}& 3 (1.5) & 5 (2) & 1 (0) \\
\hline
\multicolumn{4}{l}{$^{\mathrm{*}}$Preference: Don't disturb anyone watching a football match} \\
\multicolumn{4}{l}{$^{\mathrm{**}}$The human is sitting on the bed, watching TV.} \\
\end{tabular}
\label{tab:tab1}
\caption{This table shows the costs and clearance of the nodes bed,human, and armchair for the trajectory showed in Fig. \ref{fig:images/pipeline1.pdf}, given the human position and preferences}
\end{center}
\end{table}

In this scenario, the human is sitting on the bed and watching TV. Notably, without considering activity-based relationships, the armchair is assigned an unnecessarily high cost, despite no one occupying it. This occurs because, without relational context, the system cannot infer that the human is already seated elsewhere. When relations are incorporated, the cost of the armchair is minimized, as the system recognizes that the human is sitting on the bed instead. Although the human's cost decreases from 10 to 5 when relations are considered, it remains high, with the same clearance value, meaning that the planner’s behavior would be unaffected.

\section{Conclusion}
In this work, we explored how augmenting 3D scene graphs of static scenes (without dynamic agents) plays a key role in formulating plans that are socially aware of human behaviors and produce results that would otherwise be impossible without human-centered considerations. To evaluate this, we manually generated new 3D scene graph structures, placing humans in coherent relationships with objects in the scene, and observed how an LLM-based planner adapts under these conditions. In future extensions of this work, we aim to integrate the computed costs into a planner to generate trajectories and evaluate their effectiveness. Additionally, we plan to conduct a user study to assess the acceptability of these trajectories by comparing our approach with baseline methods. Participants will evaluate the generated trajectories in context, providing insights into human preferences and the perceived social appropriateness of different planning strategies. This approach lays the foundation for integrating 3D scene graph relationships into planning, enabling more context-aware and socially intelligent robot navigation.

\bibliographystyle{unsrt}
\bibliography{main}

\end{document}